%% file: main.tex
\documentclass{article}
\usepackage{iclr2027_conference,times}
\input{preamble}
\usepackage{hyperref}
\hypersetup{hidelinks,pdfauthor={Xiangqun Zhang, Likai Wang, Zekun Qian, Rui Zhou, Ruize Han, Wei Feng},pdftitle={CRTrack-Cap: A New Benchmark for Cross-View Dense Video Object Captioning}}
\usepackage{url}

\title{CRTRACK-CAP: A NEW BENCHMARK\\
FOR CROSS-VIEW DENSE VIDEO\\
OBJECT CAPTIONING}

\author{%
Xiangqun Zhang\textsuperscript{1}\quad
Likai Wang\textsuperscript{1}\quad
Zekun Qian\textsuperscript{1}\\[2pt]
\textbf{Rui Zhou\textsuperscript{1}}\quad
\textbf{Ruize Han\textsuperscript{2}\thanks{Corresponding author.}}\quad
\textbf{Wei Feng\textsuperscript{1}}\\[5pt]
\textnormal{\small\textsuperscript{1}School of Computer Science and Technology,}\\
\textnormal{\small Tianjin University, Tianjin, China}\\[2pt]
\textnormal{\small\textsuperscript{2}Faculty of Computer Science and Artificial Intelligence,}\\
\textnormal{\small Shenzhen University of Advanced Technology, Shenzhen, China}
}
\iclrfinalcopy

\begin{document}
\maketitle


\input{sec/0_abstract}

\input{sec/1_introduction}
\input{sec/2_related_work}
\input{sec/3_benchmark}
\input{sec/3_method}
\input{sec/4_experiments}
\input{sec/6_conclusion}
\label{sec:main_text_end}

\clearpage
\input{sec/7_submission_statements}

\bibliography{references}
\bibliographystyle{iclr2027_conference}

\clearpage
\appendix
\input{sec/S_diagnostics}

\end{document}

%% file: preamble.tex
\usepackage{graphicx}
\usepackage{amsmath,amssymb,amsfonts}
\usepackage{booktabs}
\usepackage{multirow}
\usepackage{tabularx}
\usepackage{makecell}
\usepackage{xcolor}
\usepackage{microtype}
\usepackage{enumitem}
\usepackage{xspace}
\usepackage{placeins}
\usepackage{flafter}
\usepackage{float}

\newcommand{\cvidf}{\mathrm{CVIDF1}}
\newcommand{\cvhota}{\mathrm{CV\mbox{-}HOTA}}
\newcommand{\cvassa}{\text{CV-AssA}}

\newcommand{\capa}{\mathrm{Global\mbox{-}CapA}}

\newcommand{\R}{\mathbb{R}}

%% file: sec/0_abstract.tex
\begin{abstract}
Existing dense video object captioning (DVOC) methods track and describe objects within individual videos, leaving trajectories and descriptions disconnected across cameras and complementary visual evidence unused. We introduce cross-view DVOC (CVDVOC), a more general task that associates observations of the same object across synchronized cameras into a global trajectory and generates a unified description from multiple views. To support systematic study of this task, we construct CRTrack-Cap, a new dataset containing 425 global trajectories and 838 camera-local tracklets across 13 scenes, with reference appearance descriptions for 335 identities. We also establish a unified evaluation protocol that jointly assesses global tracking and caption quality under scene-disjoint in-domain and cross-domain settings. We further propose an evidence-guided framework that associates trajectories of the same object across cameras and selectively integrates complementary visual evidence to generate a unified description. For each appearance attribute, the framework selects informative observations and regulates their contributions according to evidence reliability and cross-view disagreement. This allows complementary details to enrich global descriptions while limiting the influence of uncertain or conflicting information. Experiments demonstrate superior caption quality and joint tracking and captioning performance over adapted single-view baselines on both in-domain and cross-domain tests, establishing a strong baseline for CVDVOC.
\end{abstract}

%% file: sec/1_introduction.tex
\section{Introduction}
\label{sec:intro}

Multi-object tracking (MOT) localizes objects and maintains their identities over time, providing a foundation for understanding dynamic scenes~\citep{zhang2022bytetrack,zeng2022motr}. Dense video object captioning (DVOC) extends this capability by describing individual object trajectories in natural language, connecting object localization and identity association with semantic understanding~\citep{yang2025densevoc,wang2024semanticmot,fiastre2026captionformer}. Such descriptions make tracked objects easier to interpret and retrieve, supporting a broad range of applications, e.g., video search, human-robot interaction, and assistive scene understanding.

\begin{figure*}[t]
  \centering
  \includegraphics[width=.9\textwidth]{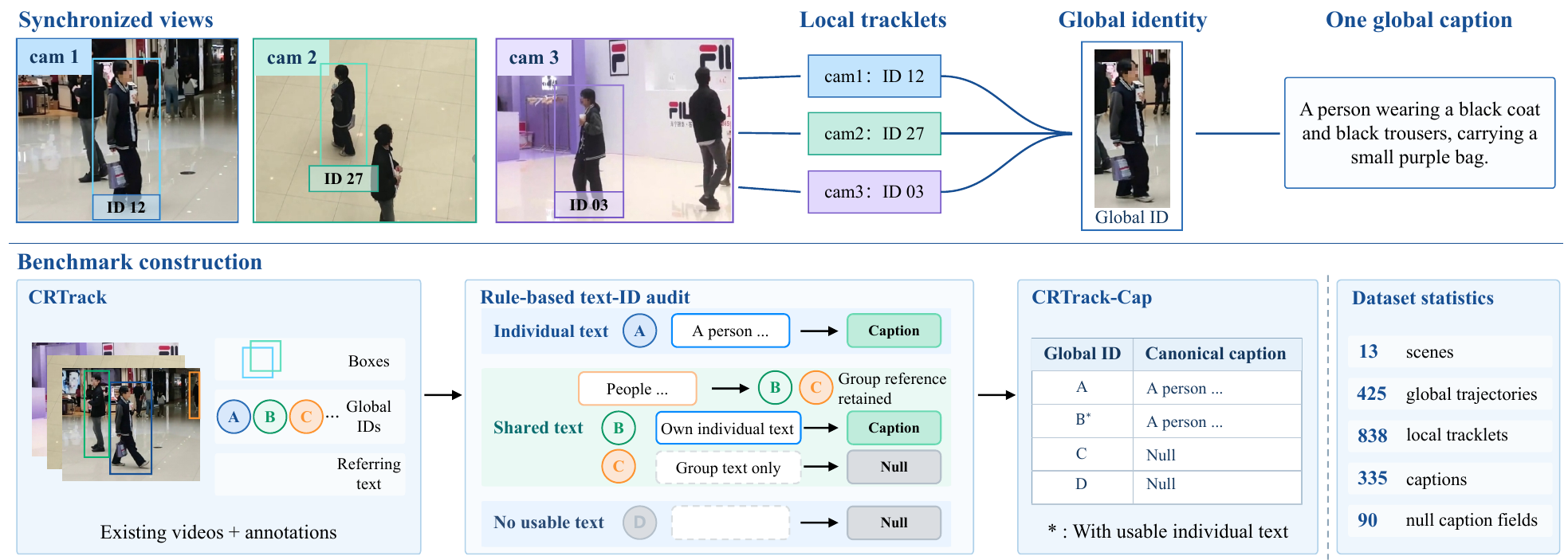}
    \caption{Cross-view DVOC and construction of CRTrack-Cap. Top: synchronized observations with different camera-local identifiers are linked into one global trajectory with one caption. Bottom: CRTrack videos, boxes, global identities, and referring expressions are converted into an auditable caption layer. Individual text supports a canonical caption for A and B; shared text retains its group scope and does not by itself supply a caption for C. D has no usable text. C and D retain their tracking annotations with null canonical-caption fields.}
  \label{fig:overview}
\end{figure*}

Existing DVOC methods are primarily designed for individual videos. When applied independently to synchronized camera views, they produce separate trajectories and descriptions for the same object, leaving its identity and semantic information disconnected across cameras. Moreover, individual views often provide incomplete appearance evidence because of occlusion, viewing direction, and image quality. For example, a pedestrian's backpack may be hidden in a frontal view but clearly visible from another camera. These observations suggest that describing an object across cameras requires both associating its observations and integrating the complementary information they provide.

To address this gap, we introduce \textbf{Cross-View Dense Video Object Captioning (CVDVOC)}, a more general task that extends DVOC from individual videos to synchronized camera views. Given multiple videos of a scene, CVDVOC aims to localize and track objects, associate their camera-local trajectories into global identities, and generate one unified description for each global trajectory. Here, a global trajectory comprises the local trajectories associated with the same object across cameras. The task therefore requires consistent identity association across views and effective use of complementary visual evidence for description generation. Figure~\ref{fig:overview} illustrates this extension.

A primary obstacle to studying CVDVOC is the lack of a benchmark that combines synchronized videos, cross-camera identities, and reference descriptions for global trajectories. Existing resources provide different parts of this requirement: BenSMOT supplies trajectory-level descriptions within individual videos~\citep{wang2024semanticmot}, while DIVOTrack focuses on cross-camera identity correspondence~\citep{wang2024divotrack}. Referring tracking data additionally link language to cross-camera targets~\citep{chen2025crmot}, but these expressions are intended to select objects and cannot be directly treated as individual trajectory captions. One expression may refer to several people, several expressions may describe the same person, and some identities have no usable individual description. Constructing a suitable dataset therefore requires establishing reliable correspondences between global identities and their descriptions.

To overcome these difficulties, we construct \textbf{CRTrack-Cap} through an annotation-conversion pipeline that audits the correspondence between referring expressions and global identities. The pipeline distinguishes individual descriptions from shared references and consolidates compatible individual expressions into canonical captions focused on stable appearance attributes. Identities without usable individual descriptions retain their tracking annotations, preserving their role in tracking evaluation while keeping caption supervision and scoring well defined. CRTrack-Cap contains 425 global trajectories formed from 838 camera-local tracklets across 13 scenes, with reference descriptions for 335 identities. We further establish a unified evaluation protocol that jointly assesses object localization, global identity association, and caption quality on scene-disjoint in-domain and cross-domain tests.

Building on this benchmark, we propose an \textbf{evidence-guided framework} that associates trajectories of the same object across cameras and selectively integrates complementary visual evidence to generate a unified description. View-Exclusive Cross-Camera Association (VECA) forms global identity groups while enforcing at most one local tracklet from each camera in a group. For description generation, Conflict-Evidence Gating (CEG) separates the selection of informative visual evidence from the control of its influence on the caption. It selects contributing views separately for each appearance attribute, evaluates reliable visual support and cross-view disagreement, and uses the resulting gated evidence to guide caption generation. This design allows different views to contribute complementary appearance details while limiting the influence of uncertain or conflicting information.

Experiments on CRTrack-Cap demonstrate that our framework outperforms adapted single-view baselines in caption quality and joint tracking and captioning performance on both in-domain and cross-domain tests. Component comparisons further show that cross-camera association improves identity consistency, while selective evidence fusion provides additional caption gains with the associated trajectories fixed. These results establish a strong baseline for CVDVOC and demonstrate the value of explicitly controlling how multi-view evidence contributes to global descriptions.

Our main contributions are summarized as follows:
\begin{itemize}
    \item We introduce \textbf{CVDVOC}, a more general problem that remains unaddressed by existing single-view DVOC methods. It aims to associate object trajectories across synchronized cameras and generate a unified description for each global trajectory using complementary visual observations.

    \item We construct \textbf{CRTrack-Cap} through an annotation-conversion pipeline that establishes reliable links between global identities and trajectory descriptions. The dataset contains 425 global trajectories across 13 scenes, with reference descriptions for 335 identities, and supports joint tracking and caption evaluation under scene-disjoint in-domain and cross-domain protocols.

    \item We propose an \textbf{evidence-guided framework} that combines view-exclusive identity association with attribute-level evidence selection and conflict control for global caption generation. The framework achieves superior caption quality and joint task performance over adapted single-view baselines on both evaluation settings.
\end{itemize}

%% file: sec/2_related_work.tex
\section{Related Work}
\label{sec:related}

\paragraph{Dense Video Object Captioning.}
Dense video captioning localizes and describes events~\citep{krishna2017denseevents,wang2021pdvc,yang2023vid2seq}, with recent extensions to streaming inference and event-duration imbalance~\citep{wang2024streamingdvc,lee2025eventequalized}. Region-level approaches connect localization with language through dense-caption pretraining and region-to-text generation, as in CapDet and GRiT~\citep{long2023capdet,wu2024grit}; video localized narratives and grounded pretraining extend this connection over time~\citep{voigtlaender2023videolocalized,yang2025groundedpretrain}. Complementary video-grounding methods, including STVGBert and TubeDETR~\citep{su2021stvgbert,yang2022tubedetr}, localize trajectories specified by input language. These tasks differ in whether language specifies or describes a target.

DVOC makes the object trajectory the description unit. DenseVOC jointly detects, tracks, and captions objects from disjoint supervision~\citep{yang2025densevoc}; BenSMOT provides semantic tracking annotations, and SMOTer predicts trajectories and their descriptions~\citep{wang2024semanticmot}. OW-VISCapTor extends instance segmentation with open-world captioning~\citep{fiastre2024owviscaptor}, and CaptionFormer unifies segmentation, tracking, and trajectory captioning~\citep{fiastre2026captionformer}. CRTrack-Cap extends this camera-local unit to global trajectories, combining observations of one identity across cameras.

\vspace{-6pt}

\paragraph{Cross-View Multi-Object Tracking.}
Single-camera MOT supplies local trajectories and appearance features for cross-view processing. Detection-based methods include ByteTrack and UCMCTrack~\citep{zhang2022bytetrack,yi2024ucmctrack}; FairMOT and RelationTrack jointly learn detection and appearance~\citep{zhang2021fairmot,yu2022relationtrack}. Transformer-based methods model temporal relations or propagate track queries, as in Global Tracking Transformers, MOTR, and MOTRv2~\citep{zhou2022gtr,zeng2022motr,zhou2023motr2}. Recent work explores identity prediction, learnable association, and discriminative embedding refinement~\citep{gao2024motip,wang2025lamotr,shao2026fdta}. These methods provide the camera-local inputs that must be associated before caption evidence can be combined across views.

Across cameras, temporal continuity alone no longer resolves identity. Deep assignment networks model relations between observations~\citep{han2022deepassignment}. MvMHAT uses self-supervised multi-view association~\citep{han2021selfsupervised}, and MvMHAT++ further studies spatial--temporal consistency and cross-view evaluation~\citep{feng2025mvmhat}. DIVOTrack, All-Day MCMT, and GMT develop cross-view benchmarks or association systems under diverse conditions~\citep{wang2024divotrack,fan2025admcmt,zhen2026gmt}; MITracker studies multi-view integration in the related visual object tracking setting~\citep{xu2025mitracker}. Appearance representations face additional variation from clothing changes and synthetic-to-real transfer, as studied in video person Re-ID~\citep{wang2025clotheschange,zhang2025syntheticrealreid}.

Language introduces a further distinction. Open-vocabulary tracking broadens the tracked category set, with COVTrack integrating semantic, motion, and appearance cues~\citep{qian2025covtrack}. Referring tracking instead uses language to select targets: iKUN connects text to existing trackers~\citep{du2024ikun}, CRMOT extends this problem across cameras~\citep{chen2025crmot}, and CD-RMOT-Bench studies cross-domain transfer~\citep{zhang2026cdrmotbench}. Cross-view DVOC instead extends caption generation across cameras, using association to collect observations of the same target.

\vspace{-6pt}

\paragraph{Evaluation of Tracking and Captioning.}
Tracking benchmarks assess localization and identity continuity~\citep{dendorfer2021motchallenge,ristani2016multicamera}; HOTA separates detection and association accuracy~\citep{luiten2021hota}, and multi-view evaluation additionally measures cross-camera identity consistency~\citep{feng2025mvmhat,chen2025crmot}. Caption metrics assess different aspects of language agreement: CIDEr measures consensus, METEOR lexical matching, and SPICE semantic structure~\citep{vedantam2015cider,banerjee2005meteor,anderson2016spice}. DenseVOC introduced CHOTA to couple detection, association, and caption accuracy, and CaptionFormer adopted it for DVOC~\citep{yang2025densevoc,fiastre2026captionformer}. We evaluate global trajectories with CV-CHOTA and its tracking and language components.

%% file: sec/3_benchmark.tex
\section{CRTrack-Cap Benchmark}
\label{sec:benchmark}

\subsection{Task and Prediction Unit}

Given synchronized videos $\mathcal V=\{V^k\}_{k=1}^{K}$, cross-view DVOC predicts global trajectories and one caption per trajectory. Our framework first extracts camera-local tracklets with identifiers, frame indices, and boxes. Each tracklet is augmented with two features that serve different downstream roles: a CaptionFormer trajectory query for language generation and an appearance descriptor for association. The resulting common record is
\begin{equation}
\mathcal T=\{T_i^k\},\qquad
T_i^k=(k,\ell_i^k,\mathcal F_i^k,\mathcal B_i^k,
\mathbf q_i^k,\mathbf e_i^k).
\label{eq:localtracklet}
\end{equation}
Here $i$ indexes tracklets within camera $k$; $\ell_i^k$ is the local identifier, $\mathcal F_i^k$ is the set of observed frames, and $\mathcal B_i^k$ contains boxes in $(x_{\min},y_{\min},x_{\max},y_{\max})$ format. A local tracklet is a camera-specific temporal unit before cross-view association. The query $\mathbf q_i^k\in\R^{d_q}$ carries the visual representation consumed by the caption pathway, whereas $\mathbf e_i^k\in\R^{d_e}$ represents appearance for cross-camera matching.

A cross-view system partitions the local tracklets into global identity groups $\widehat{\mathcal G}=\{\widehat G_n\}$. A scene-qualified identity and its time-stamped member tracklets constitute a global trajectory; the members retain their native camera coordinates. The system predicts one global caption $\widehat C_n$ per group:
\begin{equation}
\widehat G_n\subseteq\mathcal T,\qquad
\widehat Y_n=(\widehat G_n,\widehat C_n).
\label{eq:globaloutput}
\end{equation}
Here $n$ indexes predicted groups, which are disjoint and cover $\mathcal T$; $\widehat Y_n$ is the global-trajectory-and-caption prediction. A ground-truth key $g=(\mathrm{scene},\mathrm{id})$ distinguishes identities across scenes. A missing reference caption is denoted by $c_g=\varnothing$; the identity remains in tracking evaluation but is excluded from caption supervision.

\subsection{An Auditable Caption Layer}

CRTrack supplies referring expressions rather than a one-caption-per-global-trajectory annotation. Let $\mathcal E$ denote its expressions and $\mathcal I$ the scene-qualified ground-truth global identities. We represent its supervision as a relation $R\subseteq\mathcal E\times\mathcal I$ and retain both directions:
\begin{equation}
\Phi_E(e)=\{g:(e,g)\in R\},\qquad
\Phi_G(g)=\{e:(e,g)\in R\}.
\label{eq:textmap}
\end{equation}
The two relation maps expose four cases: unique mappings, multiple expressions for one identity, expressions shared by several identities, and identities without text. Figure~\ref{fig:overview} illustrates their treatment. Unique candidates undergo light normalization that preserves visible attributes. Compatible alternatives are retained as references, with one selected as the canonical caption. Conflicting descriptions remain traceable through the review record. Shared expressions retain their group scope: B's additional individual text supports its canonical caption, whereas C's shared reference alone does not. Identities without usable individual language receive a null canonical field and retain their local tracklets and global identity.

Converted samples retain source expressions, relation maps, coordinates, and frame paths. CRTrack-Cap uses these records to evaluate cross-view identity association and captions of stable, visible appearance, while reusing the released boxes and cross-camera identities.

\subsection{Dataset Composition and Splits}

We use scene-disjoint training, validation, in-domain test, and cross-domain test splits based on CRTrack~\citep{chen2025crmot} (Appendix~\ref{sec:split_details}). Training and in-domain test scenes come from DIVOTrack's moving-camera videos; cross-domain test scenes come from CAMPUS's static-camera videos. The latter therefore evaluate transfer across source domains and capture conditions. Tracking evaluation retains all identities, while caption supervision and scoring use those with canonical descriptions. Table~\ref{tab:dataset} reports both populations.

\begin{table}[!htbp]
\centering
\caption{CRTrack-Cap statistics. Coverage is the percentage of global trajectories with a canonical caption.}
\label{tab:dataset}
\footnotesize
\setlength{\tabcolsep}{4.2pt}
\renewcommand{\arraystretch}{0.96}
\begin{tabular}{lccccc}
\toprule
Split & Scenes & Identities & Captioned & Missing & Coverage\\
\midrule
Train & 6 & 256 & 196 & 60 & 76.56\\
Validation & 1 & 33 & 14 & 19 & 42.42\\
In-domain test & 3 & 87 & 76 & 11 & 87.36\\
Cross-domain test & 3 & 49 & 49 & 0 & 100.00\\
\midrule
Total & 13 & 425 & 335 & 90 & 78.82\\
\bottomrule
\end{tabular}
\vspace{-0.45em}
\end{table}

The text--identity audit identifies 341 one-text/one-identity candidates, three multiple-text/one-identity cases, and 81 identities without a source expression. Nine additional candidates lack usable canonical text. These counts preserve every global identity while defining the 335-identity language subset. Serialization, coordinate, and local-to-global checks are summarized in Appendix~\ref{sec:supp_diagnostics}.

\subsection{Scene-level Tracking and Caption Evaluation}
\label{sec:metrics}

Synchronized camera sequences form a scene-level pseudo-sequence with global identity labels. TrackEval computes CVIDF1 (global identity F1), CV-AssA (association accuracy), and CV-HOTA (joint detection and association accuracy) on this representation~\citep{feng2025mvmhat}. CV-HOTA and CV-AssA average over the standard intersection-over-union threshold grid; the caption-aware joint metric uses detection and association accuracy at threshold $0.50$. Appendix~\ref{sec:metric_details} defines the tracking metrics.

For language evaluation, Hungarian matching on HOTA's identity-alignment matrix pairs predicted groups with ground-truth identities. For identity $g$, let $m(g)$ denote its matched group, if any, and define $\hat c_g=\widehat C_{m(g)}$ when matched and an empty string otherwise. Each caption-eligible identity contributes once; unmatched predicted groups affect only tracking terms. Following CaptionFormer's scorer combination, we define
\begin{equation}
\begin{aligned}
\capa=\frac{1}{3|\mathcal G_{\mathrm{cap}}|}
\sum_{g\in\mathcal G_{\mathrm{cap}}}\bigl[&\mathrm{CIDEr}_{\mathcal D}(c_g,\hat c_g)
+\mathrm{METEOR}_{\mathcal D}(c_g,\hat c_g)\\[-2pt]
&+\mathrm{SPICE}_{\mathcal D}(c_g,\hat c_g)\bigr].
\end{aligned}
\label{eq:globalcapa}
\end{equation}
Here $\mathcal G_{\mathrm{cap}}$ contains all scene-qualified ground-truth identities with a canonical caption $c_g$, and $\hat c_g$ is the aligned prediction (empty if unaligned). Each scorer processes the full indexed pair list $\mathcal D=((c_g,\hat c_g))_{g\in\mathcal G_{\mathrm{cap}}}$ once; the subscript denotes its returned per-identity scores, including CIDEr's corpus-dependent statistics~\citep{vedantam2015cider}.

The joint score combines detection, global association, and caption accuracy:
\begin{equation}
\mathrm{CV\mbox{-}CHOTA}_{0.50}
=\bigl(\mathrm{DetA}_{0.50}\cdot\cvassa_{0.50}\cdot\capa\bigr)^{1/3}.
\label{eq:cvchota}
\end{equation}

All caption components retain their native pycocoevalcap scales. CIDEr can exceed one; consequently, Global-CapA and CV-CHOTA need not lie in $[0,1]$. In tables, CV-CHOTA uses the $0.50$ threshold and all metrics are displayed on the $\times100$ scale. Reporting the components separates language from association changes.

%% file: sec/3_method.tex
\section{Reference Framework}
\label{sec:method}

\begin{figure}[!t]
  \centering
  \includegraphics[width=.94\linewidth]{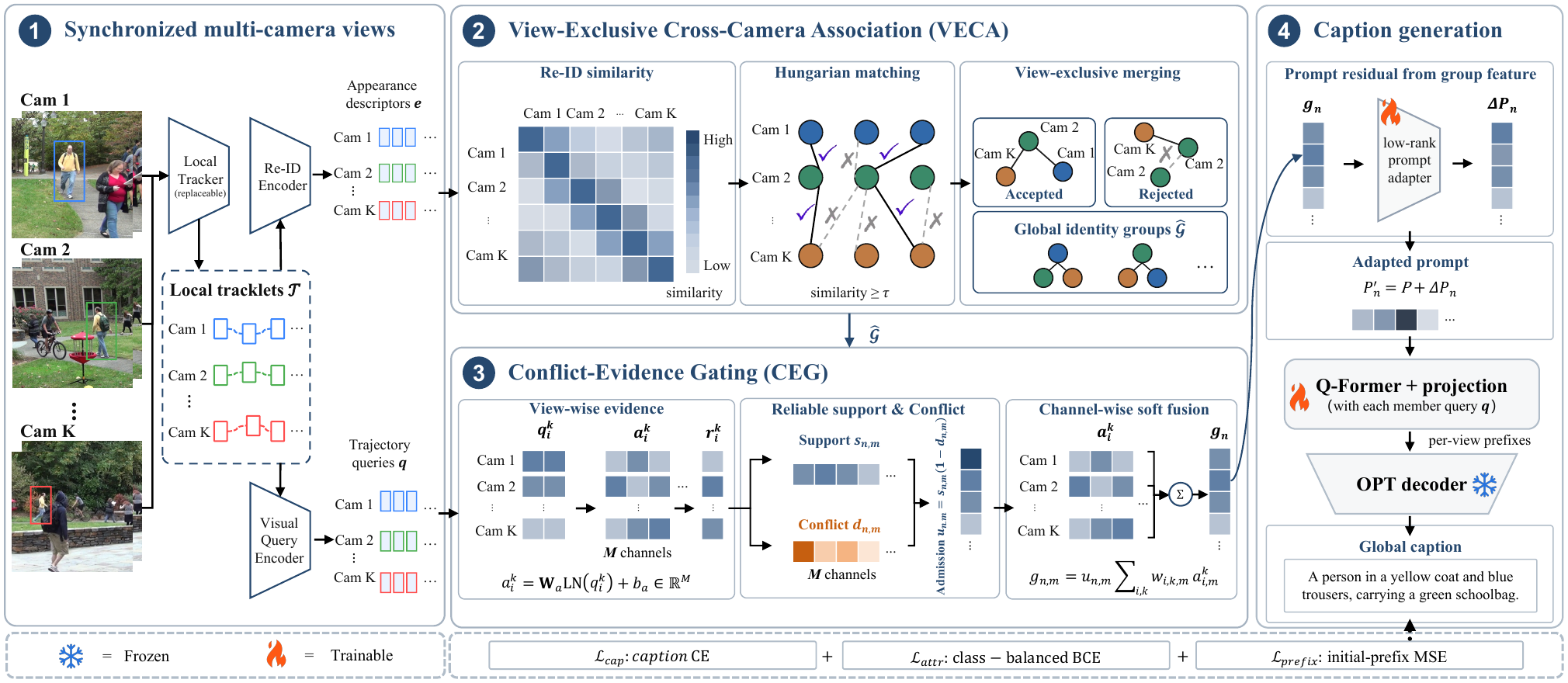}
  \caption{Reference framework for cross-view DVOC. The local tracker exports tracklets, Re-ID descriptors $\mathbf e$, and trajectory queries $\mathbf q$. VECA forms global identity groups; CEG combines member evidence with reliability and conflict estimates to produce a prompt residual $\Delta P_n$. The Q-Former processes each member query with this adapted prompt, and projected prefixes are averaged before frozen OPT generates one caption. We use CRTracker as the local tracker. The bottom row shows the training losses in Equation~\ref{eq:training}.}
  \label{fig:modules}
\end{figure}

Our framework follows the task decomposition in Figure~\ref{fig:modules}. A local front end exports tracklet geometry, an appearance descriptor for cross-camera association, and a CaptionFormer query for language generation. VECA forms view-exclusive global identity groups, and CEG uses query-derived support and disagreement to adapt their caption prompts.

\subsection{View-Exclusive Cross-Camera Association}
\label{sec:veca}

VECA addresses identity fragmentation over the fixed local tracklets. For tracklets in different cameras, it normalizes appearance descriptors and computes their similarity:
\begin{equation}
\bar{\mathbf e}_i^k=\frac{\mathbf e_i^k}{\|\mathbf e_i^k\|_2},\qquad
S_{ij}^{ku}=\langle\bar{\mathbf e}_i^k,\bar{\mathbf e}_j^u\rangle,
\qquad k\neq u.
\label{eq:cosine}
\end{equation}
For each unordered camera pair, Hungarian assignment maximizes total similarity. Assigned edges are retained when $S_{ij}^{ku}\geq\tau$, with $\tau=0.54$ calibrated by pairwise association F1 on training annotations. Pairwise optimality is insufficient for a valid global identity group because transitive links can place two tracklets from one camera in the same component. VECA therefore processes retained edges in descending score order and merges components $A$ and $B$ only if
\begin{equation}
\mathcal C(A)\cap\mathcal C(B)=\varnothing.
\label{eq:viewexclusive}
\end{equation}
Here $\mathcal C(A)$ is the set of cameras represented in $A$. Tracklets without accepted links remain singleton global identity groups. This merge enforces VECA's one-member-per-camera assumption, $|\widehat G_n\cap\mathcal T^k|\leq1$, where $\mathcal T^k$ contains the tracklets from camera $k$. VECA returns both group membership and the original member tracklets, allowing the caption pathway to inspect every contributing view.

\subsection{Conflict-Evidence Gating}
\label{sec:ceg}

For each member query in a VECA group, CEG uses a scalar reliability from alignment and tracklet geometry:
\begin{equation}
r_i^k=\max(J_i^k,0.05)\sqrt{A_i^k}
\min\!\left(1,\frac{\log(1+L_i^k)}{\log 21}\right).
\label{eq:reliability}
\end{equation}
Here $J_i^k\in[0,1]$ is the cached mask--tracklet-box IoU, $A_i^k\in[10^{-6},1]$ is representative box area normalized by the padded, resized image area and clipped to that interval, and $L_i^k$ is tracklet length. Larger aligned observations receive more weight, while duration support saturates at 20 frames. This geometric visibility proxy is computed independently of the semantic logits; the same scalar accompanies all evidence channels of that tracklet.

MVMean averages member language-prefix matrices after the query adapter, Q-Former, and projection. We denote this complete member pathway by $\operatorname{CFPrefix}(\mathbf q;P)$, with learned prompt $P$ (Appendix~\ref{sec:reliability_details}).

CEG weights members separately for each evidence channel. A channel head maps each normalized member query into logits:
\begin{equation}
\mathbf a_i^k=\mathbf W_a\,\operatorname{LN}(\mathbf q_i^k)+\mathbf b_a,
\qquad \mathbf a_i^k\in\R^M,\quad \mathbf W_a\in\R^{M\times d_q}.
\label{eq:attrlogits}
\end{equation}
The $M=88$ channels provide structured weak supervision over stable appearance attributes. Each channel predicts a weak label obtained by deterministic parsing of the canonical training captions. An unmentioned attribute is coded as zero for this supervision, although silence in a caption does not establish visual absence. A class-balanced binary cross-entropy term with weight $0.05$ supervises the fused group logits before the admission gate (Appendix~\ref{sec:reliability_details}). For channel $m$, let $p_{i,m}^k=\operatorname{sigmoid}(a_{i,m}^k)$ and $\chi_{i,m}^k=|2p_{i,m}^k-1|$. The first quantity predicts the caption-derived label, while the second measures confidence away from the ambiguous midpoint.

Let $\mathcal I_n=\{(i,k):T_i^k\in\widehat G_n\}$ index group members. All sums, maxima, and softmax operations below run over $\mathcal I_n$, separately for each channel $m$. The normalized reliabilities are
\begin{equation}
\pi_i^k=\frac{r_i^k}{\sum_{(j,v)\in\mathcal I_n}r_j^v},
\qquad
\hat r_i^k=\frac{r_i^k}{\max_{(j,v)\in\mathcal I_n}r_j^v}.
\label{eq:normalized_reliability}
\end{equation} With the training-set reliability percentile $r_{0.9}$ as an absolute-quality reference, support and conflict are
\begin{equation}
\begin{aligned}
s_{n,m}&=\operatorname{clip}\!\left(\frac{\max_{(i,k)\in\mathcal I_n}r_i^k}{r_{0.9}},0,1\right)
\max_{(i,k)\in\mathcal I_n}\bigl(\hat r_i^k \chi_{i,m}^k\bigr),\\[-2pt]
d_{n,m}&=\operatorname{clip}\!\left(
4\sum_{(i,k)\in\mathcal I_n}\pi_i^k(p_{i,m}^k-\bar p_{n,m})^2,0,1\right),\\[-2pt]
\bar p_{n,m}&=\sum_{(i,k)\in\mathcal I_n}\pi_i^k p_{i,m}^k.
\end{aligned}
\label{eq:conflict}
\end{equation}
Support measures reliable certainty; conflict measures probability disagreement. Their product $u_{n,m}=s_{n,m}(1-d_{n,m})$ admits evidence that is both supported and cross-view coherent. CEG then performs channel-specific soft fusion:
\begin{equation}
\begin{aligned}
w_{i,k,m}&=\operatorname{softmax}_{(i,k)\in\mathcal I_n}
\left(\log\max(r_i^k,\epsilon)+a_{i,m}^k\right),\\[-2pt]
z_{n,m}&=\sum_{(i,k)\in\mathcal I_n}w_{i,k,m}a_{i,m}^k,
\qquad g_{n,m}=u_{n,m}z_{n,m}.
\end{aligned}
\label{eq:filtered_evidence}
\end{equation}
Here $\epsilon$ is the floating-point machine epsilon used to clamp reliability before taking the logarithm. The softmax weights select the members that contribute to channel $m$, while $u_{n,m}$ determines how strongly their fused evidence can modify the caption prompt. This separates evidence selection from conflict suppression.

A rank-$8$ adapter maps $\mathbf g_n\in\R^M$ to a residual $\Delta P_n\in\R^{N_q\times d_p}$ for the $N_q$ learned Q-Former tokens of width $d_p$:
\begin{equation}
\begin{aligned}
P_n'&=P+\Delta P_n,\\
\mathbf p_n&=\frac{1}{|\mathcal I_n|}\sum_{(i,k)\in\mathcal I_n}
\operatorname{CFPrefix}(\mathbf q_i^k;P_n').
\end{aligned}
\label{eq:prompt}
\end{equation}
Here $\rho_n$ is the mean admission across channels, and $0.1\rho_n$ scales the residual (Appendix~\ref{sec:reliability_details}). Members share the adapted prompt; their averaged prefix conditions the frozen OPT decoder~\citep{zhang2022opt}. The query adapter, Q-Former (including $P$), and projections form the 107.36M trainable caption pathway; CEG adds 245,200 parameters. The visual encoder, tracker, Re-ID encoder, and OPT remain frozen, and VECA is fixed.

\subsection{Training Objective}

The caption pathway learns to generate the canonical description while retaining the pretrained prefix representation:
\begin{equation}
\mathcal L=\mathcal L_{\mathrm{cap}}
+0.05\,\mathcal L_{\mathrm{attr}}
+\mathcal L_{\mathrm{prefix}}.
\label{eq:training}
\end{equation}
Here $\mathcal L_{\mathrm{cap}}$ is teacher-forced caption cross-entropy, $\mathcal L_{\mathrm{attr}}$ is class-balanced binary cross-entropy on the pre-admission logits $z_{n,m}$ with caption-derived weak labels, and $\mathcal L_{\mathrm{prefix}}$ is mean-squared error between the group prefix $\mathbf p_n$ and its initial CaptionFormer counterpart. The weak-label term grounds evidence channels in the training captions; the prefix term anchors adaptation to the pretrained language interface. Appendix~\ref{sec:reliability_details} specifies optimization details.

%% file: sec/4_experiments.tex
\section{Experiments}
\label{sec:experiments}

\subsection{Experimental Setup}

\paragraph{Comparison protocol.}
To our knowledge, no existing method directly addresses cross-view dense video object captioning (CVDVOC). We therefore adapt three single-view comparison methods capable of trajectory-level description---DenseVOC~\citep{yang2025densevoc}, SMOTer~\citep{wang2024semanticmot}, and CaptionFormer~\citep{fiastre2026captionformer}. After fine-tuning their native tracking components, we apply the same MvMHAT~\citep{han2021selfsupervised} algorithm to associate their local trajectories across synchronized views. We evaluate the resulting global trajectories and captions on the same CRTrack-Cap splits with a common evaluator. Our framework uses VECA for cross-view association and CEG for caption fusion; Table~\ref{tab:ablation} compares the component configurations on fixed local predictions.

\paragraph{Training and evaluation.}
For caption training, local tracklets and their assigned global identity groups are fixed. We optimize the 107.36M-parameter caption pathway and the 245K-parameter CEG adapter, using one caption-eligible global identity per step. Training runs for 70 epochs with AdamW, learning rates of $10^{-6}$ and $10^{-4}$ for the caption pathway and CEG, respectively, weight decay $10^{-2}$, and gradient clipping at one. We report full-split results on the in-domain and cross-domain tests.

\subsection{Scene-level Task Performance}

\begin{table}[!htbp]
\centering
\caption{Main comparison on CRTrack-Cap with predicted local tracklets. DenseVOC, SMOTer, and CaptionFormer use MvMHAT for cross-view association; ours uses VECA. Metric values are shown on the $\times100$ scale; higher is better.}
\label{tab:main}
\setlength{\tabcolsep}{4.2pt}
\resizebox{0.88\linewidth}{!}{%
\begin{tabular}{lcccccc}
\toprule
Method & CVIDF1 & CV-HOTA & CV-AssA & CV-CHOTA & Global-CapA & CIDEr\\
\midrule
\multicolumn{7}{l}{\textit{In-domain test}}\\
DenseVOC & 66.28 & 55.79 & 51.64 & 48.17 & 24.63 & 26.70\\
SMOTer & 59.83 & 51.80 & 44.82 & 57.43 & 48.18 & 66.39\\
CaptionFormer & 77.35 & 64.28 & 58.44 & 76.34 & 75.54 & 133.53\\
\textbf{Ours} & \textbf{80.65} & \textbf{68.57} & \textbf{62.84} & \textbf{83.43} & \textbf{88.19} & \textbf{157.04}\\
\midrule
\multicolumn{7}{l}{\textit{Cross-domain test}}\\
DenseVOC & 15.69 & 16.15 & 5.98 & 19.58 & 19.11 & 17.14\\
SMOTer & \textbf{18.87} & \textbf{20.88} & \textbf{8.22} & 30.82 & 41.45 & 63.36\\
CaptionFormer & 16.40 & 16.87 & 6.38 & 29.94 & 60.63 & 97.70\\
\textbf{Ours} & 17.00 & 18.71 & 7.27 & \textbf{33.48} & \textbf{68.34} & \textbf{112.62}\\
\bottomrule
\end{tabular}}
\end{table}

Table~\ref{tab:main} compares complete pipelines with different local trackers, caption pathways, and association rules. Our framework leads all in-domain metrics. Cross-domain tracking is strongest for SMOTer, but our framework leads CV-CHOTA, Global-CapA, and CIDEr, showing that tracking strength alone does not determine global-caption quality. The component study separates association from caption fusion.

\subsection{Separating Association from Caption Fusion}
\label{sec:ablation}

Table~\ref{tab:ablation} compares a $2\times2$ grid using the same predicted local tracklets and evaluator. The association configurations are singleton groups and VECA groups; the caption configurations use separately trained MVMean and CEG pathways. The VECA+CEG rows report the main-table results.

\begin{table}[!htbp]
\centering
\caption{Association and caption-fusion configurations; metric values are shown on the $\times100$ scale. The VECA+CEG rows reproduce Table~\ref{tab:main}; the other rows are the recorded component controls.}
\label{tab:ablation}
\setlength{\tabcolsep}{3.7pt}
\renewcommand{\arraystretch}{0.92}
\resizebox{\linewidth}{!}{%
\begin{tabular}{lcccccccc}
\toprule
Setting & VECA & CEG & CVIDF1 & CV-HOTA & CV-AssA & CV-CHOTA & Global-CapA & CIDEr\\
\midrule
\multicolumn{9}{l}{\textit{In-domain test}}\\
Singleton + MVMean & & & 50.65 & 51.86 & 35.96 & 67.04 & 80.19 & 138.03\\
Singleton + CEG & & $\checkmark$ & 50.65 & 51.86 & 35.96 & 68.56 & 85.77 & 149.01\\
VECA + MVMean & $\checkmark$ & & \textbf{80.65} & \textbf{68.57} & \textbf{62.84} & 79.92 & 77.52 & 130.20\\
VECA + CEG & $\checkmark$ & $\checkmark$ & \textbf{80.65} & \textbf{68.57} & \textbf{62.84} & \textbf{83.43} & \textbf{88.19} & \textbf{157.04}\\
\midrule
\multicolumn{9}{l}{\textit{Cross-domain test}}\\
Singleton + MVMean & & & 14.87 & 17.34 & 6.25 & 29.83 & 56.37 & 87.86\\
Singleton + CEG & & $\checkmark$ & 14.87 & 17.34 & 6.25 & 29.54 & 54.73 & 82.67\\
VECA + MVMean & $\checkmark$ & & \textbf{17.00} & \textbf{18.71} & \textbf{7.27} & 31.45 & 56.64 & 82.46\\
VECA + CEG & $\checkmark$ & $\checkmark$ & \textbf{17.00} & \textbf{18.71} & \textbf{7.27} & \textbf{33.48} & \textbf{68.34} & \textbf{112.62}\\
\bottomrule
\end{tabular}}
\end{table}

VECA raises in-domain CVIDF1 by 30.00 points and CV-AssA by 26.88 over singleton groups, but MVMean lowers Global-CapA after this grouping change. With VECA fixed, CEG raises Global-CapA from 77.52 to 88.19 in-domain and from 56.64 to 68.34 cross-domain, while tracking metrics remain unchanged. Thus, association supplies the global groups, whereas evidence-aware fusion determines how those groups improve captions. CEG lowers caption scores for singleton groups but improves them with VECA groups fixed, and VECA+CEG attains the grid's highest joint and caption scores on both splits.

The recorded association-rule comparison gives 80.65 in-domain CVIDF1 for full VECA versus 78.22 for greedy view-exclusive matching. Unconstrained graph closure creates 5 in-domain and 46 cross-domain same-camera collisions; view-exclusive groups avoid these invalid links (Appendix~\ref{sec:association_ablation_supp}).

\paragraph{View-count analysis.}
Table~\ref{tab:view_count_main} stratifies the VECA+CEG and VECA+MVMean configurations from Table~\ref{tab:ablation} by the number of cameras observing each caption-eligible reference identity.


\begin{table}[!htbp]
\centering
\caption{View-count comparison with VECA groups fixed. Metric deltas are CEG minus MVMean on the $\times100$ scale. Sample counts and additional metrics appear in Appendix~\ref{sec:view_count_supp}.}
\label{tab:view_count_main}
\setlength{\tabcolsep}{8pt}
\renewcommand{\arraystretch}{0.84}
\begin{tabular}{lccc}
\toprule
Split & Views & $\Delta$CV-CHOTA & $\Delta$CIDEr\\
\midrule
In-domain & 1 & +0.11 & +0.89\\
In-domain & 2 & +6.55 & +55.03\\
In-domain & 3 & +4.11 & +38.48\\
Cross-domain & 3 & +2.22 & +35.21\\
Cross-domain & 4 & +1.88 & +29.13\\
\bottomrule
\end{tabular}
\end{table}

The single-view in-domain group changes little, whereas both multi-view groups improve in CIDEr and CV-CHOTA; the two cross-domain groups also gain. This pattern supports the role of CEG in combining evidence across views, while the non-monotonic gains do not imply that adding cameras always yields larger improvements. Together, these results indicate that cross-view caption quality depends on both correct grouping and selective use of complementary evidence.

\subsection{Qualitative Comparison}

\begin{figure}[!htb]
  \centering
  \includegraphics[width=1\linewidth]{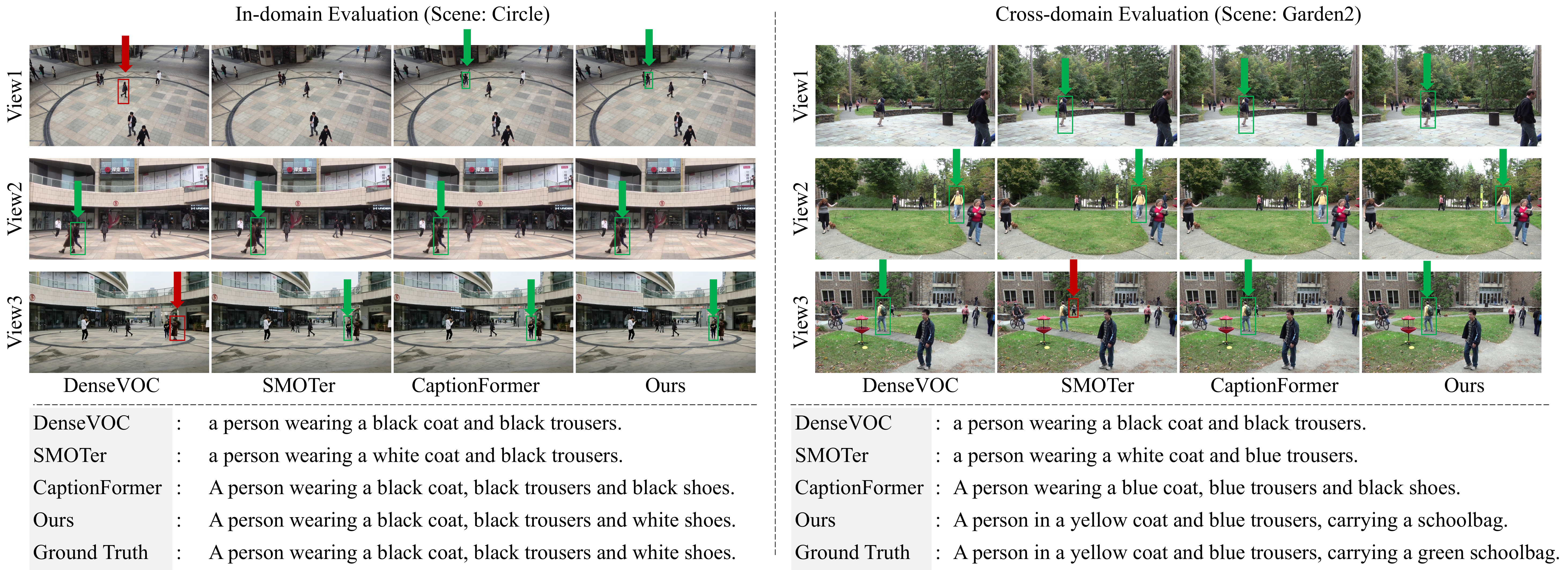}
  \caption{In-domain (Circle) and cross-domain (Garden2) examples. Rows are synchronized views; columns are methods. Green/red arrows mark correct/incorrect links; no arrow means unmatched. Text below compares generated captions with ground truth.}
  \label{fig:mechanism_main}
  \vspace{-0.5em}
\end{figure}

Figure~\ref{fig:mechanism_main} compares in-domain and cross-domain targets. Our framework and CaptionFormer link all shown views, but our captions better match the reference attributes: white shoes in the former and a yellow coat and schoolbag in the latter.

\FloatBarrier

%% file: sec/6_conclusion.tex
\section{Conclusion}
\label{sec:conclusion}


We extended single-view DVOC to captioning global trajectories across cameras. CRTrack-Cap supports this task with 425 tracking identities and 335 canonical captions, audited from referring expressions and evaluated on scene-disjoint in-domain and cross-domain tests. Our reference framework combines cross-view association with caption fusion and achieves the highest CV-CHOTA among the compared methods on both tests. Component comparisons distinguish association and caption gains under fixed local predictions. These results support evaluating association and caption quality jointly. They also show that the caption gains depend on how the associated evidence is fused, rather than on grouping alone. Cross-domain performance remains lower, motivating further study of captioning across capture conditions.

%% file: sec/7_submission_statements.tex




\subsection*{AI use statement}
Generative AI tools were used to polish the manuscript’s language and improve its readability. All AI-assisted content was reviewed and verified by the authors. The authors take full responsibility for the final content of the paper.

%% file: sec/S_diagnostics.tex
\section{Additional Analyses and Reproducibility Details}
\label{sec:supp_diagnostics}

This appendix follows the evaluated pipeline: caption-layer construction, cross-view association, CEG implementation, caption-evidence diagnostics, and metric computation.

\subsection{Benchmark Construction and Source Audit}

The index retains all 425 scene-qualified global trajectories and 838 local tracklets, including the 90 identities with null captions. Of the 425 identity records, 341 are one-text/one-identity candidates, three contain multiple expressions for one identity, and 81 have no source expression. Nine additional candidates lack usable canonical text. The release therefore contains a 425-identity tracking index and a 335-identity language-eligible subset.

Source validation records 4,936 border-clipped boxes, 11 sequence-dimension mismatches, and 138 annotation rows with unavailable frame images. These diagnostics are retained alongside source coordinates and image paths. Converted samples serialize without errors, and the local-to-global mapping check reports no conflicts. Conversion validity and source-frame availability are recorded as separate fields, preserving all missing-frame outcomes for audit.

\subsection{Scene Splits and Domain Shift}
\label{sec:split_details}

The split manifest uses Floor, Gate1, Ground, Moving, Park, and Shop for training; Square for validation; Circle, Gate2, and Side for in-domain testing; and Garden1, Garden2, and ParkingLot for cross-domain testing. We retain CRMOT's in-domain and cross-domain test scenes and hold Square out from its original training scenes for development.

The domain distinction follows the source collections in CRTrack~\citep{chen2025crmot}. Training and in-domain scenes originate from DIVOTrack, whereas the cross-domain scenes originate from CAMPUS. DIVOTrack uses three moving cameras; CAMPUS uses three or four static cameras. The change in capture setup, scene appearance, viewpoint, and trajectory duration makes cross-domain evaluation a transfer across source domains rather than only a new scene from the training collection.

\subsection{Association-Rule Analysis}
\label{sec:association_ablation_supp}

Table~\ref{tab:association_control} compares the recorded association-rule settings on the in-domain test. Full VECA scores 80.65/68.57/62.84 CVIDF1/CV-HOTA/CV-AssA, compared with 78.22/67.75/61.18 for greedy view-exclusive matching. Their CV-DetA scores are close (74.99 versus 75.02). Ordinary graph closure produces 5 in-domain and 46 cross-domain same-camera collisions; view-exclusive merging avoids these invalid groups.

\begin{table}[htbp]
\centering
\caption{Recorded association-rule comparison on the in-domain test. TrackEval metric values are shown on the $\times100$ scale; best values are bold.}
\label{tab:association_control}
\footnotesize
\setlength{\tabcolsep}{5pt}
\renewcommand{\arraystretch}{0.92}
\resizebox{\linewidth}{!}{%
\begin{tabular}{lccccc}
\toprule
Rule & CVIDF1 & CV-HOTA & CV-DetA & CV-AssA & CV-AssRe\\
\midrule
Singleton & 50.65 & 51.86 & 74.99 & 35.96 & 37.09\\
Greedy view-exclusive & 78.22 & 67.75 & \textbf{75.02} & 61.18 & 63.79\\
Hungarian star & 77.03 & 66.65 & 74.99 & 59.38 & 61.74\\
\textbf{Full VECA} & \textbf{80.65} & \textbf{68.57} & 74.99 & \textbf{62.84} & \textbf{65.39}\\
\bottomrule
\end{tabular}
}
\end{table}

For the same in-domain greedy view-exclusive run, CV-CHOTA@0.50, Global-CapA, and CIDEr are 80.00, 82.74, and 142.62, respectively.

\subsection{CEG Supervision and Optimization}
\label{sec:reliability_details}

The implementation uses $M=88$ structured weak-label channels: 22 item-type atoms and 66 slot--color atoms. The latter factor six semantic slots with eleven colors. Canonical training captions are parsed deterministically into multi-label targets. The channel logits are explicitly supervised; attribute-level caption F1 measures attributes extracted from the final decoded captions.

The geometric reliability is defined in Equation~\ref{eq:reliability}. Its reference percentile pools member reliabilities from caption-eligible training identities, with $r_{0.9}=0.08719$. Reliability supplies one scalar per local tracklet and is broadcast across the evidence channels. The query cache selects the predicted mask with maximum IoU against the supplied representative tracklet box rasterized at mask resolution; mask probabilities are thresholded at $0.5$. The box is scaled and normalized by the padded image dimensions before computing its area fraction. In the predicted-tracklet pipeline, the supplied box comes from the local tracker.

Optimization combines teacher-forced caption cross-entropy, class-balanced weak-label binary cross-entropy with weight $0.05$, and mean-squared error to the initial CaptionFormer prefix with weight $1.0$. The residual bottleneck has rank eight, its scale is $0.1$, and the up-projection is zero-initialized. The trainable caption path comprises a 33,600-parameter query adapter, approximately 105.16M Q-Former/query-token parameters, and 2,166,016 projection parameters.

Each optimization step uses one identity. Training uses no gradient accumulation, learning-rate scheduler, or warmup. The frozen OPT decoder uses half precision, while the optimizer runs without an automatic mixed-precision gradient scaler. Training uses one RTX~3090.

MVMean averages aligned language-prefix matrices from the shared query adapter, Q-Former, and projection, denoted $\operatorname{CFPrefix}(\mathbf q;P)$, where $P$ is the learned Q-Former prompt:
\begin{equation}
\mathbf p_n^{\mathrm{mv}}=
\frac{1}{|\widehat G_n|}
\sum_{T_i^k\in\widehat G_n}
\operatorname{CFPrefix}(\mathbf q_i^k;P).
\label{eq:mvmean}
\end{equation}

\paragraph{Member indices and prompt mapping.}
The pair $(i,k)\in\mathcal I_n$ identifies a tracklet, not an independent camera--tracklet Cartesian product. Several tracklets may exist in a camera before association; VECA permits at most one in each predicted group. The head uses $W_a\in\R^{M\times d_q}$ and $\mathbf b_a\in\R^M$. The query $\mathbf q_i^k$ enters the evidence head before the caption query adapter. Appearance descriptors $\mathbf e_i^k$ enter VECA only. Layer normalization acts on the feature dimension; the fusion softmax normalizes across members for each fixed channel $m$, so $\sum_{(i,k)\in\mathcal I_n}w_{i,k,m}=1$.

For $\mathbf g_n=(g_{n,1},\ldots,g_{n,M})$, the prompt adapter is
\begin{equation}
\begin{aligned}
\rho_n&=\frac{1}{M}\sum_{m=1}^{M}u_{n,m},\\
\mathbf h_n&=\operatorname{GELU}(W_{\downarrow}\operatorname{LN}(\mathbf g_n)+\mathbf b_{\downarrow}),\\
\Delta P_n&=0.1\rho_n\operatorname{reshape}_{N_q\times d_p}
\left[\tanh(W_{\uparrow}\mathbf h_n+\mathbf b_{\uparrow})\right].
\end{aligned}
\label{eq:cegresidual}
\end{equation}
Here $W_{\downarrow}\in\R^{8\times M}$, $W_{\uparrow}\in\R^{N_qd_p\times8}$, and $\mathbf h_n\in\R^8$. The up-projection weight and bias start at zero. Batch dimensions are omitted. The residual modifies only the learned prompt tokens; the projected object-query token is concatenated with them. The Q-Former output is projected into OPT's embedding space, giving a prefix with $N_q+1$ tokens. With $\operatorname{CFPrefix}$ denoting this complete member pathway, the group prefix is
\begin{equation}
\mathbf p_n=\frac{1}{|\mathcal I_n|}\sum_{(i,k)\in\mathcal I_n}
\operatorname{CFPrefix}(\mathbf q_i^k;P_n')\in\R^{(N_q+1)\times d_{\mathrm{OPT}}}.
\label{eq:cegprefix}
\end{equation}
The same fusion mapping is used in training and inference; caption targets and weak labels are used only for training. For the full CEG configuration, inference applies the residual directly, without the separate experimental sample router. Non-finite inputs, near-zero query norms, or reliability sums/maxima at most $10^{-8}$ trigger an inference fallback without a residual; training rejects such samples. The logarithm in Equation~\ref{eq:filtered_evidence} uses the dtype's machine epsilon, a distinct safeguard.

\paragraph{Weak-label loss and reduction.}
Let $N_{\mathrm{tr}}$ be the number of caption-eligible training identities and $y_{n,m}\in\{0,1\}$ the parsed target. Counts and positive-class weights are computed on those training identities:
\begin{equation}
N_m^+=\sum_{n=1}^{N_{\mathrm{tr}}}y_{n,m},\qquad
\omega_m=\operatorname{clip}\!\left(
\frac{N_{\mathrm{tr}}-N_m^+}{\max(N_m^+,1)},1,10\right).
\label{eq:positive_weights}
\end{equation}
Writing $z_{n,m}=\sum_{(i,k)\in\mathcal I_n}w_{i,k,m}a_{i,m}^k$ for the fused logits before admission, the per-identity loss is
\begin{equation}
\mathcal L_{\mathrm{attr}}(n)=-\frac{1}{M}\sum_{m=1}^{M}
\left[\omega_m y_{n,m}\log\sigma_{\mathrm{sig}}(z_{n,m})
+(1-y_{n,m})\log\!\left(1-\sigma_{\mathrm{sig}}(z_{n,m})\right)\right],
\label{eq:weak_bce}
\end{equation}
where $\sigma_{\mathrm{sig}}$ denotes sigmoid. The implementation uses numerically stable \texttt{BCEWithLogitsLoss} with \texttt{pos\_weight} and the default mean reduction over the $1\times M$ input. Positive-class weighting applies only to the positive term. Supervision acts on $z_{n,m}$, before $g_{n,m}=u_{n,m}z_{n,m}$; group labels supervise the fused representation, without separate per-view attribute annotations.

The parser encodes mentioned atoms as one and all remaining atoms as zero. These zeros represent missing caption mentions as well as genuine negatives. Accordingly, the weak-label probabilities characterize caption-derived evidence, and a confident zero is not an independently verified visual absence.

Let $\mathcal U_n$ index non-ignored, shifted caption-token targets $y_{n,t}^{\mathrm{cap}}$. Teacher forcing and prefix anchoring use
\begin{equation}
\begin{aligned}
\mathcal L_{\mathrm{cap}}(n)&=-\frac{1}{|\mathcal U_n|}\sum_{t\in\mathcal U_n}
\log p_\theta(y_{n,t}^{\mathrm{cap}}\mid y_{n,<t}^{\mathrm{cap}},\mathbf p_n),\\
\mathcal L_{\mathrm{prefix}}(n)&=\frac{\|\mathbf p_n-\mathbf p_n^{(0)}\|_F^2}{(N_q+1)d_{\mathrm{OPT}}}.
\end{aligned}
\label{eq:caption_prefix_losses}
\end{equation}
Here $\theta$ denotes the trainable caption pathway and CEG parameters. The frozen OPT decoder propagates gradients to the prefix; padding and prefix positions are ignored as targets. The cached initial prefix $\mathbf p_n^{(0)}$ is detached and stored in half precision. The prefix loss uses elementwise mean reduction with weight $1.0$; the weak-label term has weight $0.05$. One training identity contributes one optimization step.

\paragraph{Parameter accounting.}
The existing trainable caption pathway and the newly added CEG parameters are different counts. The former totals approximately 107.36M: a 33,600-parameter query adapter, approximately 105.16M Q-Former and learned-query-token parameters, and a 2,166,016-parameter language projection. CEG adds 245,200 parameters for evidence prediction and the low-rank prompt adapter. Thus the combined trainable count is approximately 107.61M. Frozen visual encoding, tracking, Re-ID, and OPT parameters are excluded from both trainable counts.

\subsection{View-Count Diagnostics}
\label{sec:view_count_supp}

This analysis stratifies the VECA+CEG and VECA+MVMean configurations in Table~\ref{tab:ablation} by reference-view count. With VECA groups fixed, the in-domain $\Delta$CIDEr is 0.89 for one-view identities and 55.03/38.48 for two-/three-view identities (Table~\ref{tab:view_count}).

\vspace{-0.4em}
\begin{table}[!htbp]
\centering
\caption{CEG--MVMean gain by reference-view count with VECA groups fixed. Metric deltas are shown on the $\times100$ scale.}
\label{tab:view_count}
\footnotesize
\setlength{\tabcolsep}{4pt}
\renewcommand{\arraystretch}{0.92}
\resizebox{\linewidth}{!}{%
\begin{tabular}{lcccccc}
\toprule
Split & Views & $\Delta$CV-CHOTA & $\Delta$Global-CapA & $\Delta$CIDEr & $\Delta$METEOR & $\Delta$SPICE\\
\midrule
In-domain & 1 & +0.11 & +0.29 & +0.89 & $-$0.02 & +0.00\\
In-domain & 2 & +6.55 & +20.06 & +55.03 & +4.05 & +1.10\\
In-domain & 3 & +4.11 & +14.17 & +38.48 & +2.76 & +1.26\\
Cross-domain & 3 & +2.22 & +12.87 & +35.21 & +2.78 & +0.62\\
Cross-domain & 4 & +1.88 & +10.75 & +29.13 & +1.96 & +1.15\\
\bottomrule
\end{tabular}
}
\end{table}

The one-view in-domain group changes little, whereas both multi-view in-domain groups improve in all three core caption-aware measures. Both cross-domain strata improve across all five reported metrics: the three- and four-view groups gain 35.21/29.13 CIDEr, 2.22/1.88 CV-CHOTA, and 12.87/10.75 Global-CapA, respectively. The two-view in-domain group has the largest CIDEr gain; the cross-domain gains in these three measures remain below those of the multi-view in-domain groups.

\subsection{Evaluation Units and Metric Definitions}
\label{sec:metric_details}

Result records link the configuration, seed, checkpoint digest, local-tracklet predictions, global identity groups, caption predictions, metric-input directory, and evaluator output. Component comparisons reuse the corresponding frozen tracklet records. Caption checkpoints are recorded per configuration, distinguishing separately trained models from inference-only interventions.

Caption evaluation uses one caption-eligible global identity as the scoring unit, irrespective of its number of frames or views. Missing reference captions remain null, and unaligned eligible references receive empty predictions. Invalid runs and unavailable source frames remain documented so that reconstruction preserves the evaluated population.

For scene-level identity assignment, $\mathrm{IDTP}$, $\mathrm{IDFP}$, and $\mathrm{IDFN}$ count identity true positives, false positives, and false negatives:

\begin{equation}
\cvidf=\frac{2\,\mathrm{IDTP}}{2\,\mathrm{IDTP}+\mathrm{IDFP}+\mathrm{IDFN}}.
\label{eq:cvidf1}
\end{equation}

At intersection-over-union threshold $\alpha$, let $\mathcal T_\alpha$ contain matched detection pairs, and let $\mathcal F_\alpha$ and $\mathcal N_\alpha$ contain unmatched predicted and reference detections. For a match $c$, $\mathrm{TPA}_\alpha(c)$, $\mathrm{FPA}_\alpha(c)$, and $\mathrm{FNA}_\alpha(c)$ count true-positive, false-positive, and false-negative associations, respectively:

\begin{equation}
\begin{aligned}
\cvassa_\alpha&=\frac{1}{|\mathcal T_\alpha|}
\sum_{c\in\mathcal T_\alpha} A_\alpha(c),\\[-2pt]
A_\alpha(c)&=\frac{\mathrm{TPA}_\alpha(c)}{\mathrm{TPA}_\alpha(c)+\mathrm{FPA}_\alpha(c)+\mathrm{FNA}_\alpha(c)}.
\end{aligned}
\label{eq:assa}
\end{equation}

Detection accuracy is
\begin{equation}
\mathrm{DetA}_\alpha=\frac{|\mathcal T_\alpha|}{|\mathcal T_\alpha|+|\mathcal F_\alpha|+|\mathcal N_\alpha|}.
\label{eq:deta}
\end{equation}
HOTA averages over $\mathcal A=\{0.05,0.10,\ldots,0.95\}$:

\begin{equation}
\cvhota=\frac{1}{|\mathcal A|}\sum_{\alpha\in\mathcal A}
\sqrt{\mathrm{DetA}_\alpha\,\cvassa_\alpha}.
\label{eq:cvhota}
\end{equation}